\documentclass{article}

\PassOptionsToPackage{numbers, compress}{natbib}

\usepackage[final]{neurips_2020}

\usepackage[utf8]{inputenc} 
\usepackage[T1]{fontenc}    
\usepackage{hyperref}       
\usepackage{url}            
\usepackage{booktabs}       
\usepackage{amsfonts}       
\usepackage{nicefrac}       
\usepackage{microtype}      
\usepackage{graphicx}
\usepackage{amsmath}
\usepackage{graphicx}
\usepackage{wrapfig}
\usepackage{booktabs}
\usepackage{float}
\usepackage{subfig}
\usepackage{amssymb}
\usepackage{algorithm}
\usepackage{algorithmic}
\usepackage{multirow}
\usepackage[frozencache,cachedir=.]{minted}

\graphicspath{{./images/}}

\title{NLPGym - A toolkit for evaluating RL agents on Natural Language Processing Tasks}

\author{%
  Rajkumar Ramamurthy\\
  Fraunhofer IAIS\\
  Sankt Augustin, Germany\\
  \texttt{rajkumar.ramamurthy@iais.fraunhofer.de} \\
  \And
  Rafet Sifa \\
  Fraunhofer IAIS \\
  Sankt Augustin, Germany \\
  \texttt{rafet.sifa@iais.fraunhofer.de} \\
  \AND
  Christian Bauckhage \\
  Fraunhofer IAIS \\
  Sankt Augustin, Germany \\
  \texttt{christian.bauckhage@iais.fraunhofer.de} \\
}

\begin{document}

\maketitle

\begin{abstract}
Reinforcement learning (RL) has recently shown impressive performance in complex game AI and robotics tasks. To a large extent, this is thanks to the availability of simulated environments such as OpenAI Gym, Atari Learning Environment, or Malmo which allow agents to learn complex tasks through interaction with virtual environments. While RL is also increasingly applied to natural language processing (NLP), there are no simulated textual environments available for researchers to apply and consistently benchmark RL on NLP tasks. With the work reported here, we therefore release NLPGym, an open-source Python toolkit that provides interactive textual environments for standard NLP tasks such as sequence tagging, multi-label classification, and question answering. We also present experimental results for 6 tasks using different RL algorithms which serve as baselines for further research. The toolkit is published at \url{https://github.com/rajcscw/nlp-gym}
\end{abstract}

\section{Introduction}

The ability to comprehend and communicate using language is a major characteristic of human intelligence. For artificial agents to be able to assist humans with language-oriented tasks, they have to acquire similar natural language understanding capabilities. Indeed, there has been substantial progress in this regard and machines have become well versed in natural language processing (NLP) tasks such as language modeling \cite{vaswani2017attention, devlin2018bert}, information extraction \cite{akbik-etal-2018-contextual, strakova-etal-2019-neural}, text summarization \cite{nallapati2016summarunner, zhong2020extractive}, or question answering \cite{hermann2015teaching, yang2019xlnet}.

 Recently, deep reinforcement learning (DRL) has become popular in NLP as most tasks can be formulated as a kind of sequence decision-making problem \cite{maes2009structured, daume2009search}.  This is promising for two reasons: First, DRL is inherently interactive and allows for training agents through human feedback in the form of rewards. 
 Second, learning based on rewards allows the agents to be trained directly based on application-specific metrics such as F1 or ROUGE scores which are generally non-differentiable and cannot be optimized easily with supervised learning.

Typical applications of DRL in NLP include: (a) sequence tagging \cite{maes2009structured, daume2009search} where RL is used to solve structured prediction tasks such as named entity recognition and part of speech tagging;  (b) text summarization \cite{ryang2012framework, gao2018april, chen-bansal-2018-fast} where agents select sentences to be included in summaries; (c) question answering \cite{wang2017r} where agents rank and select relevant paragraphs and it is interesting to note that question answering can be cast as an interactive game \cite{yuan2019interactive}; (d) information extraction \cite{narasimhan2016improving} where agents query and extract information from external resources to extract information;  (e) solving text-based games \cite{narasimhan2015language, cote2018textworld, ammanabrolu2020motivate}. 

Despite their increased use in NLP, there are no frameworks for testing RL agents on standard NLP tasks. In other words, it seems that the application of RL to NLP suffers from lack of availability of open-source frameworks which are similar to Gym \cite{brockman2016openai}, Malmo \cite{johnson2016malmo}, Arcade Learning Environment (ALE) \cite{Bellemare_2013}, TextWorld \cite{cote2018textworld}, or Baby AI \cite{chevalier2018babyai} which accelerated research on solving robotic tasks, mine-craft playing, ATARI game-playing, text-based games and ground language agents.
 
In this work, we therefore present NLPGym, a toolkit to bridge the gap between applications of RL and NLP. This aims at facilitating research and benchmarking of DRL application on natural language processing tasks. Our toolkit provides interactive environments for standard NLP tasks such as sequence tagging, question answering, and sequence classification. The environments provide standard RL interfaces and therefore can be used together with most RL frameworks such as baselines \cite{baselines}, stable-baselines \cite{stable-baselines}, and RLLib \cite{liang2018rllib}. Furthermore, the toolkit is designed in a modular fashion providing flexibility for users to extend tasks with their custom data sets, observations, and reward functions.

To illustrate the capabilities of NLPGym, this paper also provides examples and detailed experimental results on the above tasks using different RL algorithms, featurizers and reward functions. These results are intended to serve as baselines for further research in this area. 

Indeed, in future work, we plan to include environments for text summarization, text generation and machine translation tasks as they can also be formulated as MDP problems. We hope that NLPGym becomes an ever growing test-bed for testing agents for learning language and understanding.

\paragraph{Related Work} We present a short survey of related work concerning frameworks for solving language-oriented tasks.  They fall under two main categories: \textit{grounded language learning} and \textit{text-based games}. Notable works in \textit{grounded-language learning} include 3D simulated environments \cite{hermann2017grounded, chaplot2017gated} for developing language-informed agents. Building upon this idea of specifying goals using language, BabyAI \cite{chevalier2018babyai} enables agents to interact and manipulate objects in environments. In the category of \textit{text-based games}, TextWorld \cite{cote2018textworld} offers gym-style environments for solving text-based adventure games. Likewise, Jericho \cite{hausknecht2020interactive} supports a wide variety of interactive fiction games. Focussing on dialogue research, LIGHT \cite{urbanek2019learning} provides a platform for studying conversational agents that can interact with characters and objects in a large-scale text-based adventure game.  Since most toolkits only focus on language-informed tasks and text-based games, there is a lack of equivalent frameworks for developing agents to solve NLP tasks, which we address with our work of NLPGym.

\section{NLPGym Toolkit}

\begin{figure*}[t!]
    \centering
	\subfloat[Sequence Tagging with NER tags]{\includegraphics[width=0.27\textwidth]{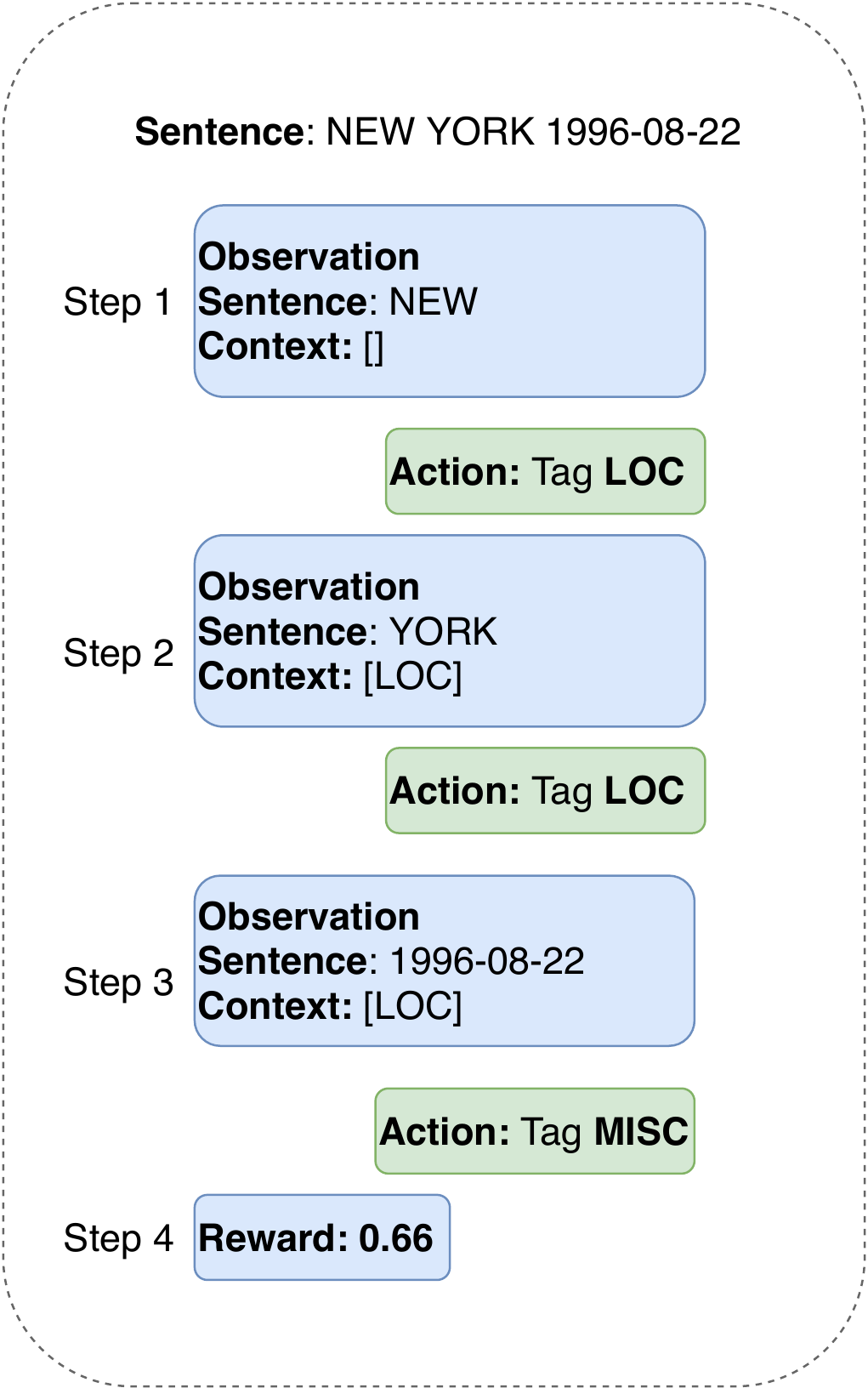}}
	\hspace{2ex}
	\subfloat[Multiple-Choice Question Answering]{\includegraphics[width=0.27\textwidth]{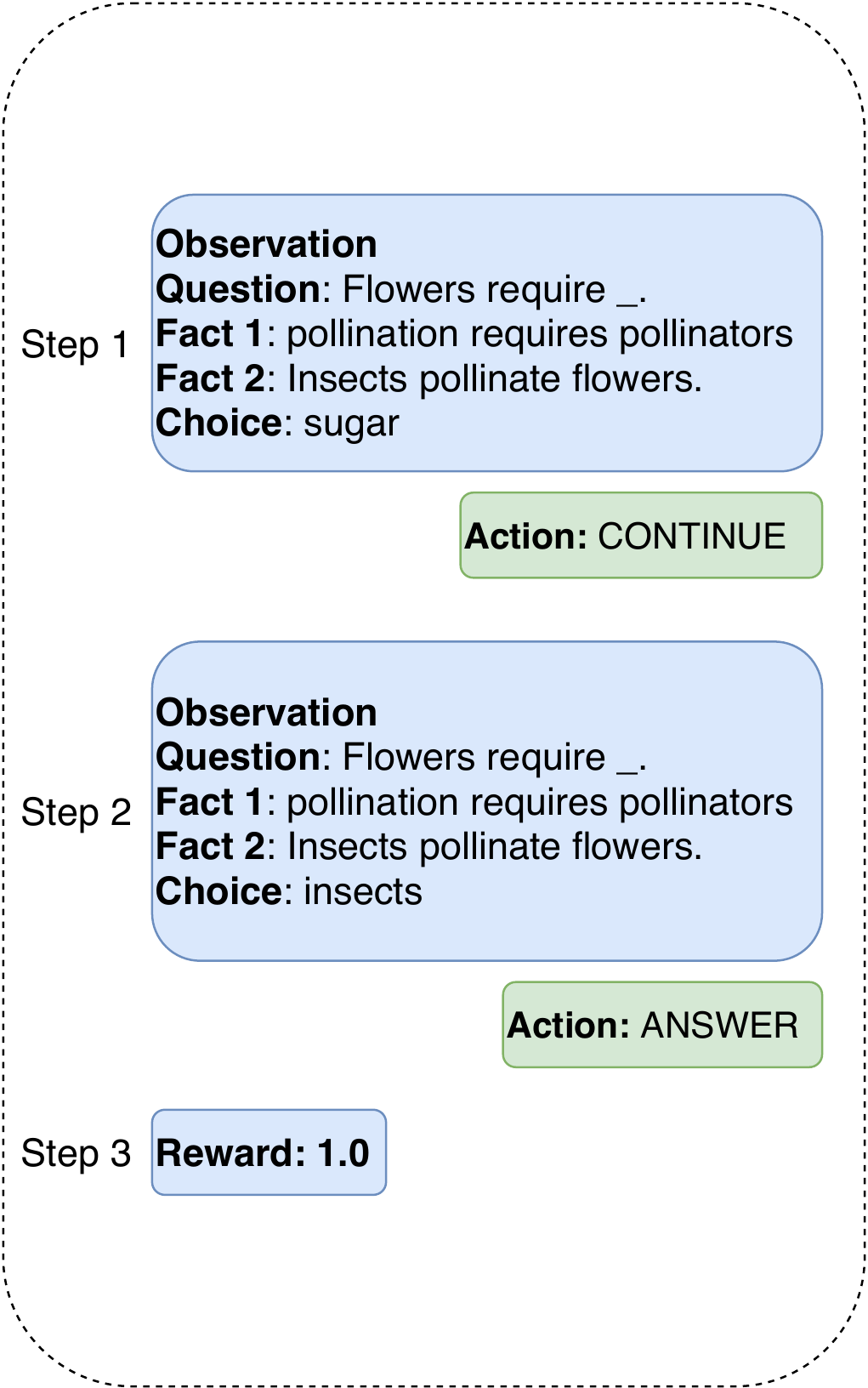}}
	\hspace{2ex}
	\subfloat[Sequence Classification]{\includegraphics[width=0.27\textwidth]{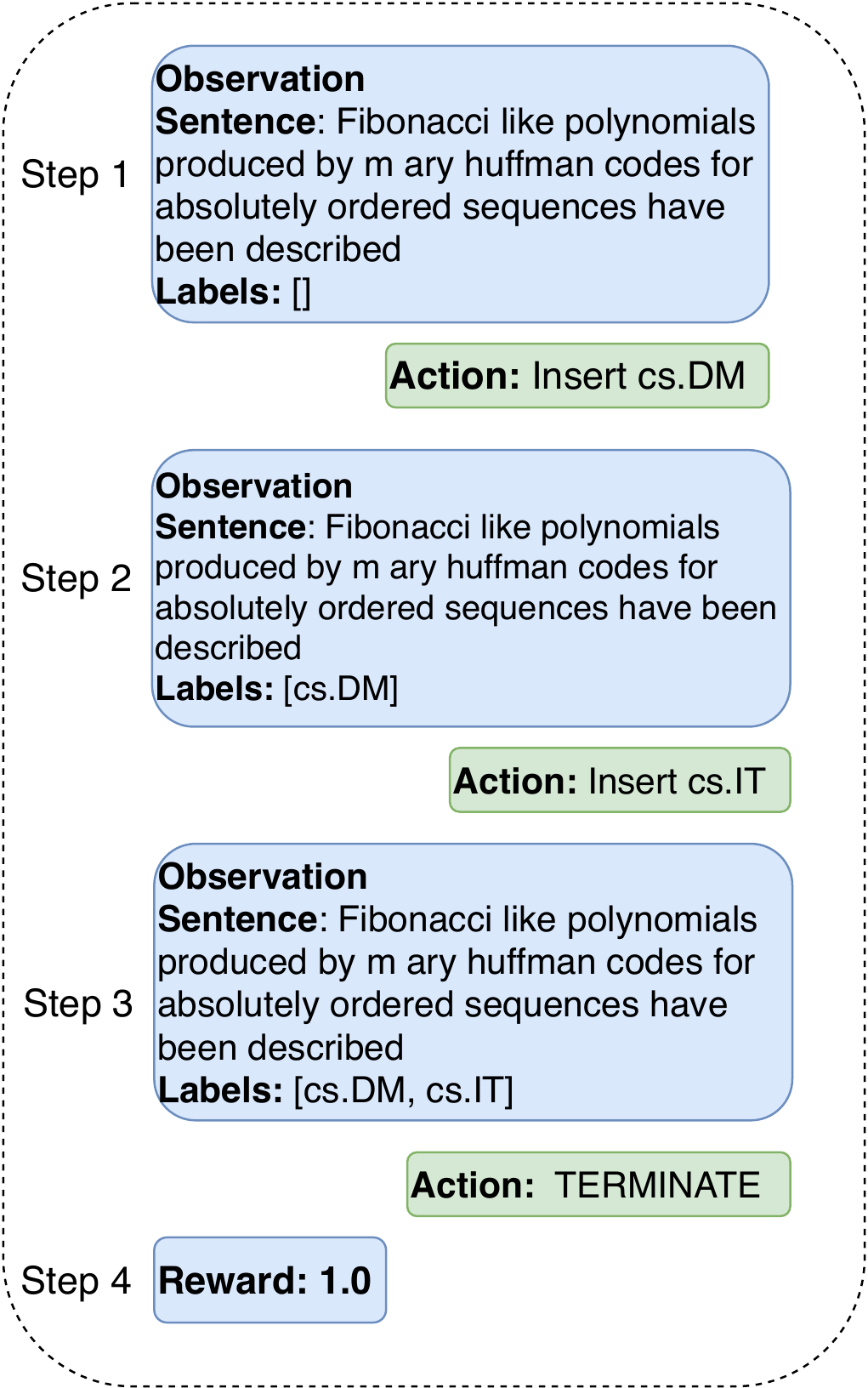}}
	\caption{\textbf{NLPGym Environments}: An overview of sample episodic interactions of agent-environments for three different tasks. (a) tagging a sequence with NER tags; (b) answering multiple-choice questions; c) generating label sequence for a given sentence}
	\label{fig:overview}
\end{figure*}

The NLPGym toolkit that we present in this work consists of interactive environments for learning natural language processing tasks.  It includes environments for three tasks: sequence tagging, question answering, and multi-label sequence classification.   Each task is formulated as a Markov Decision Process (MDP) defined by a tuple: $\langle \boldsymbol{S}, \boldsymbol{A}, \boldsymbol{T}, \boldsymbol{R} \rangle$ where $\boldsymbol{S}$ is a set of states, $\boldsymbol{A}$ is a set of actions, $\boldsymbol{R}$ is the reward function, and $\boldsymbol{T}$ is the transition probability. Figure \ref{fig:overview} shows an overview of sample episodic interactions of agent-environment for the provided tasks. 

The toolkit is implemented in Python extending on OpenAI Gym's API with \mintinline{python}{step()}, \mintinline{python}{reset()} and \mintinline{python}{render()} functions for observing and interacting with the environment.  Due to its compliance with Gym's API, NLPGym can be easily integrated with most RL frameworks \cite{baselines, stable-baselines, liang2018rllib}. A demo script to train PPO/DQN agents can be found in Appendix \ref{appendix:ppo_dqn_sample}. Moreover, the toolkit is designed in a modular fashion, allowing users to plug-in custom components such as reward functions, featurizers for observation and datasets. For more details, we refer to Appendix \ref{appendix:extensibility}, which elaborates their extensibility using code snippets. Each environment has default implementations of observation featurizers (See Appendix \ref{appendix: default}) and reward functions and therefore, can be used out-of-the-box.

\subsection{Batch vs Online Learning}

Each environment relies on annotated data points that are used in generating episodes. NLPGym supports injecting samples into the environment dynamically at any point in time. This feature offers the flexibility to develop algorithms in two settings; \textit{batch} setting in which all of the data points are added at once and an \textit{interactive/online} setting in which they are added one at a time. A demo script illustrating this has been provided in Appendix \ref{appendix: online}. Human-in-the-loop applications favor the later setting in which samples are first annotated by humans and then added to the environment. Furthermore, RL agents can pre-annotate them, thereby reducing human annotation efforts.

\subsection{Environments}

\paragraph{Sequence Tagging (ST):} Sequence tagging is one of the core tasks in NLP for natural/spoken language understanding (NLU/SLU) and typical tasks include named-entity recognition (NER) and part-of-speech (POS) tagging. Formally, given a sentence with words $(w_1, w_2, \dots w_t)$, the task is to generate its associated label sequence $(l_1, l_2, \dots l_t)$. This task can be cast as an MDP \cite{maes2009structured} in which the sentence is parsed in left-to-right order, one word at a time. The environment state $s_t$ at step $t$ consists of $w_t$ and predicted labels for previous words. However, the agent's observation $o_t$ at each step is just the current word $w_t$ and context (predicted label of the previous word $l_{t-1}$), which renders the environment partially observable. The available actions are to \textit{TAG} the current word with one of the possible labels and the action space is thus discrete.  On performing any action, the transition function is straightforward, that the tagged label is added to the label sequence. The reward is computed as the entity-level F1-score between actual and predicted labels. It comes in two flavors; sparse (given at the end of the episode) and dense (difference in scores between consecutive steps).


\paragraph{Multiple-choice Question Answering (QA):} Multiple-choice question answering (QA) is at the core of machine reading comprehension \cite{khot2019qasc, khashabi2020unifiedqa}. The task of QA is to answer a given question $q$ by selecting one of the multiple choices $c_1, c_2, \dots c_t$.  Besides, questions are often accompanied by supporting facts $f$, which contain further context.  Selecting the correct option out of all choices can be considered as a sequential decision-making task.  Each episode spawns with a question to be answered. At each step $t$, the question $q$, supporting facts $f$, and one of the multiple choices $c_t$ are given as the observation to agents. Given this observation, agents' actions are binary whether to answer \textit{ANS} or to continue with the next choice \textit{CONT}. Consequently, on choosing the action \textit{ANS}, the episode terminates and the last observed choice is considered the final answer. On the other hand, on selecting \textit{CONT}, the environment moves to the next choice and presents its corresponding observation.  The reward is given only at the end of the episode, either 0 or 1, based on the selected choice's correctness.

\paragraph{Multi-label Classification (MLC)} Multi-label classification is a generalization of several NLP tasks such as multi-class sentence classification and label ranking \cite{tsoumakas2007multi}. The task of multi-label classification is to assign a label sequence $l_1, l_2 \dots l_n$ to the given sentence. In information retrieval, this task corresponds to label ranking when preferential relation exists over labels. Likewise, the task reduces to a simple multi-class classification when any label sequence's maximum length is at most one. In any case, generating this label sequence can be cast as a sequential decision-making task. Each episode begins with a sentence and an empty label sequence.  The observation at each step $t$ is the given sentence and generated label sequence until $t$.  Similar to sequence tagging, available actions are to \textit{INSERT} one of the possible labels. Moreover, agents can terminate the episode using the \textit{TERM} action. The reward is chosen as the F1-score between actual and predicted label sequences, either sparse or dense. 


\section{Demo Experiments}\label{experiments}

\begin{figure*}[t!]
    \centering
	\subfloat[ST with NER tags]{\includegraphics[width=0.32\textwidth]{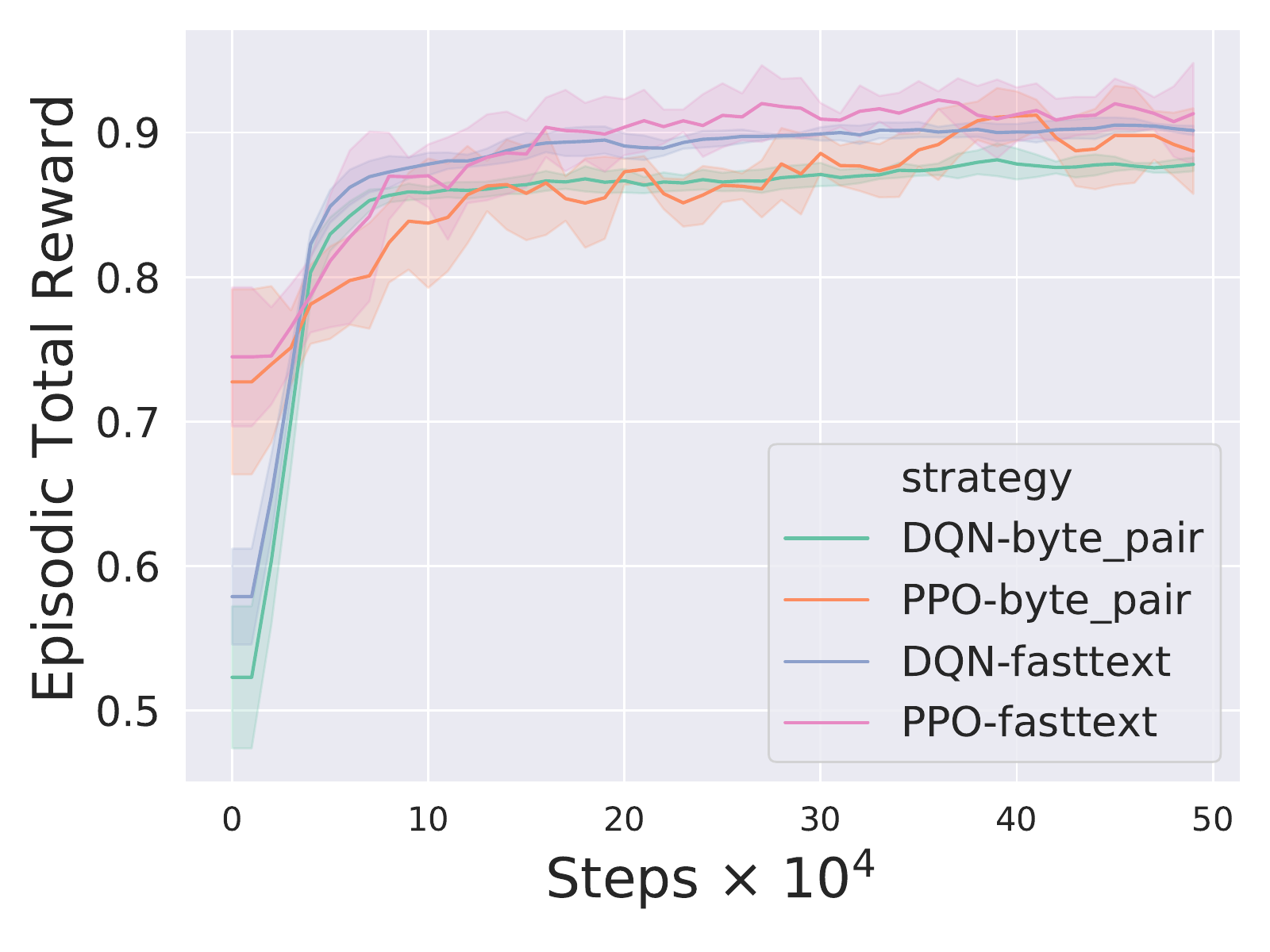}}
	\subfloat[MLC with Reuters]{\includegraphics[width=0.32\textwidth]{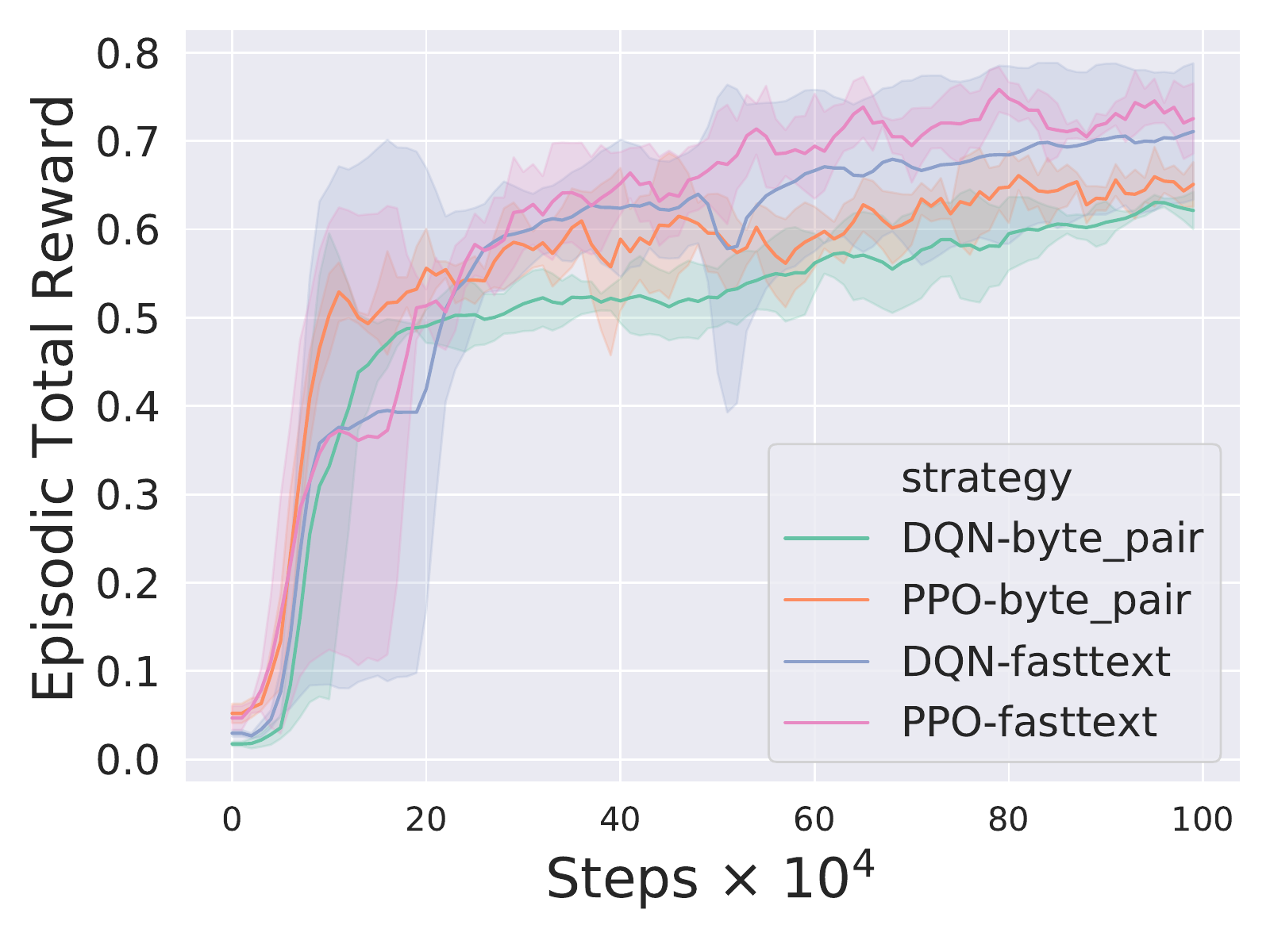}}
	\subfloat[QA with QASC]{\includegraphics[width=0.32\textwidth]{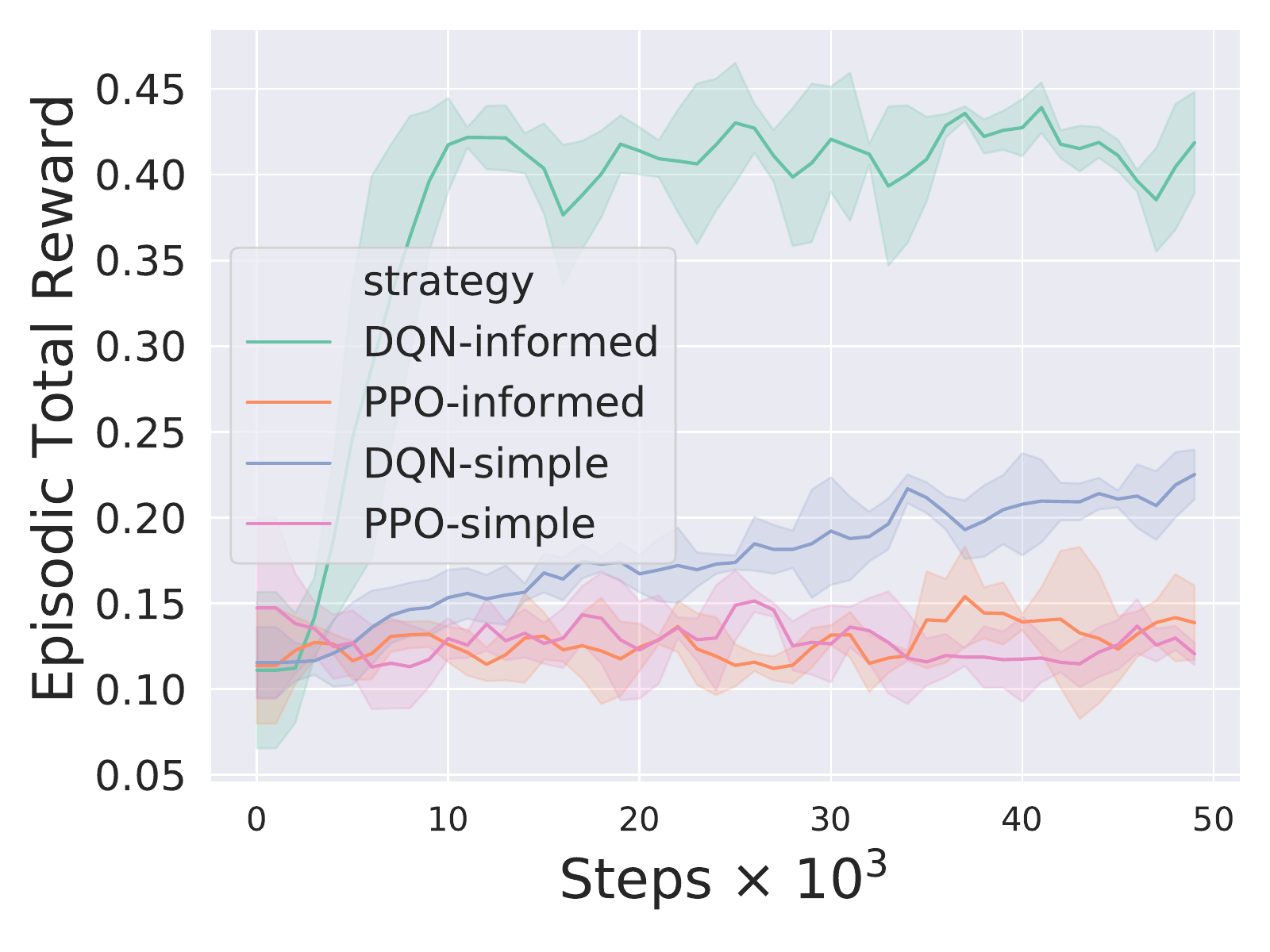}} \\
	
	\subfloat[ST with POS tags]{\includegraphics[width=0.32\textwidth]{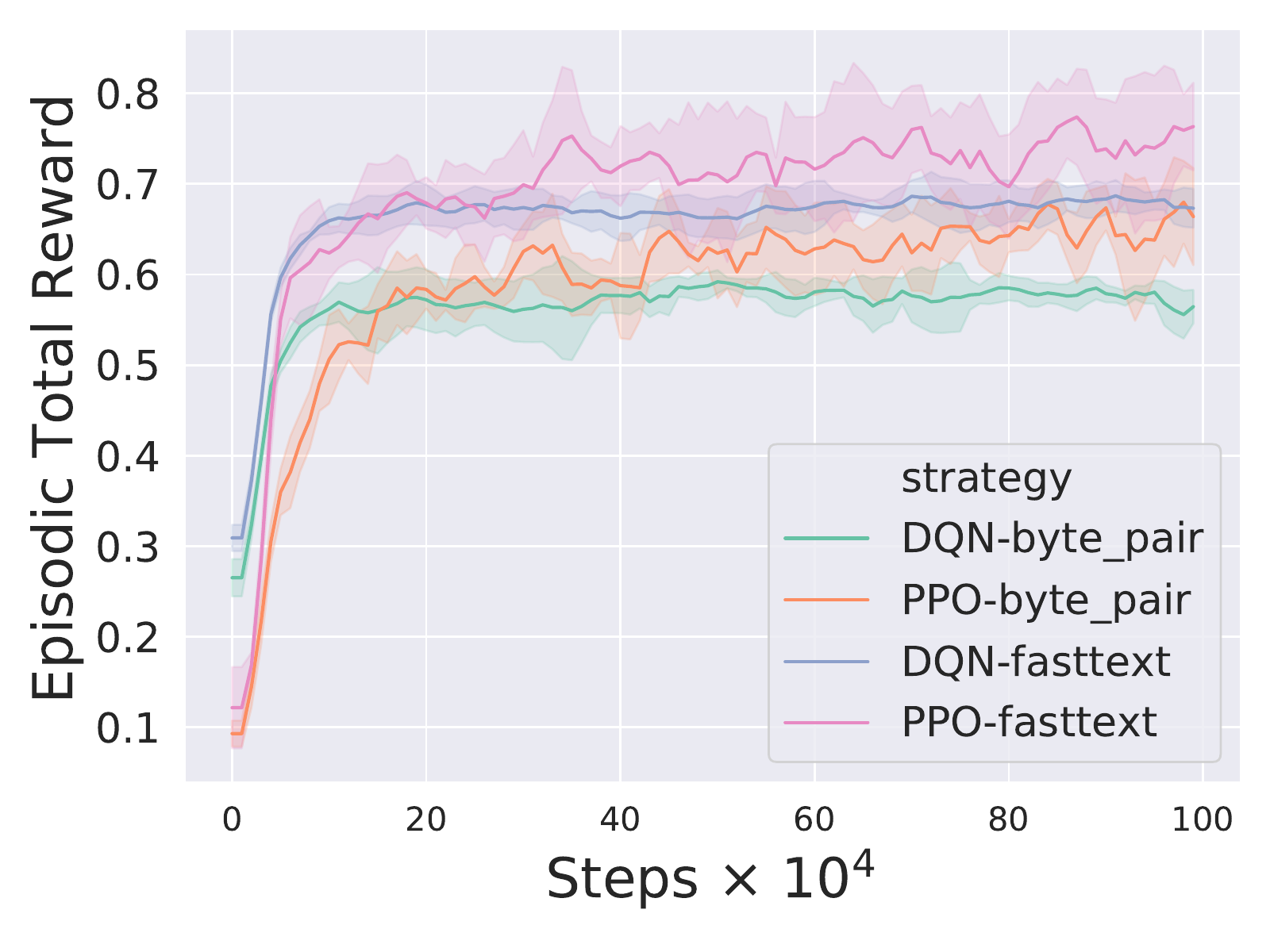}}
	\subfloat[MLC with AAPD]{\includegraphics[width=0.32\textwidth]{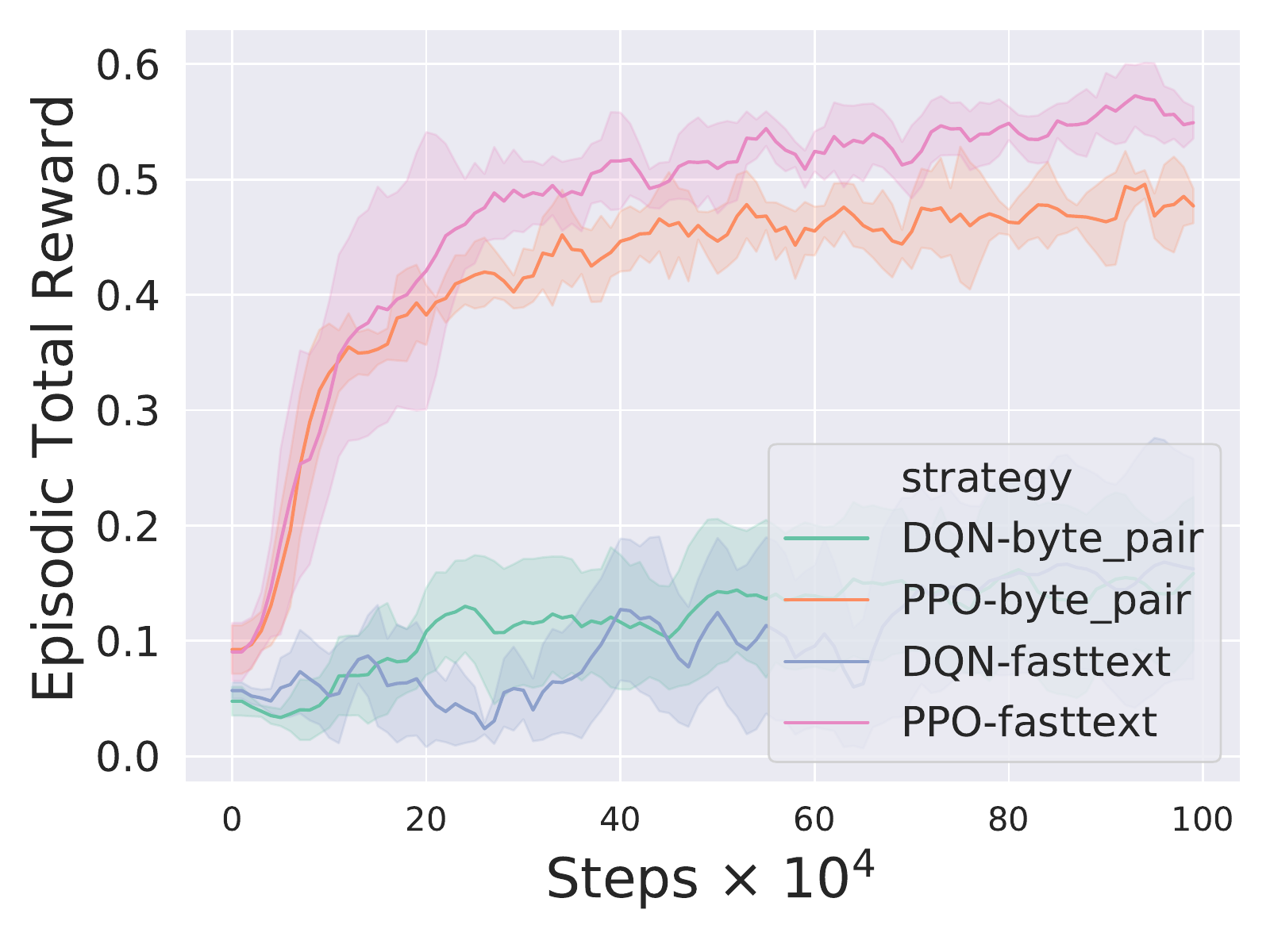}}
	\subfloat[QA with AIRC]{\includegraphics[width=0.32\textwidth]{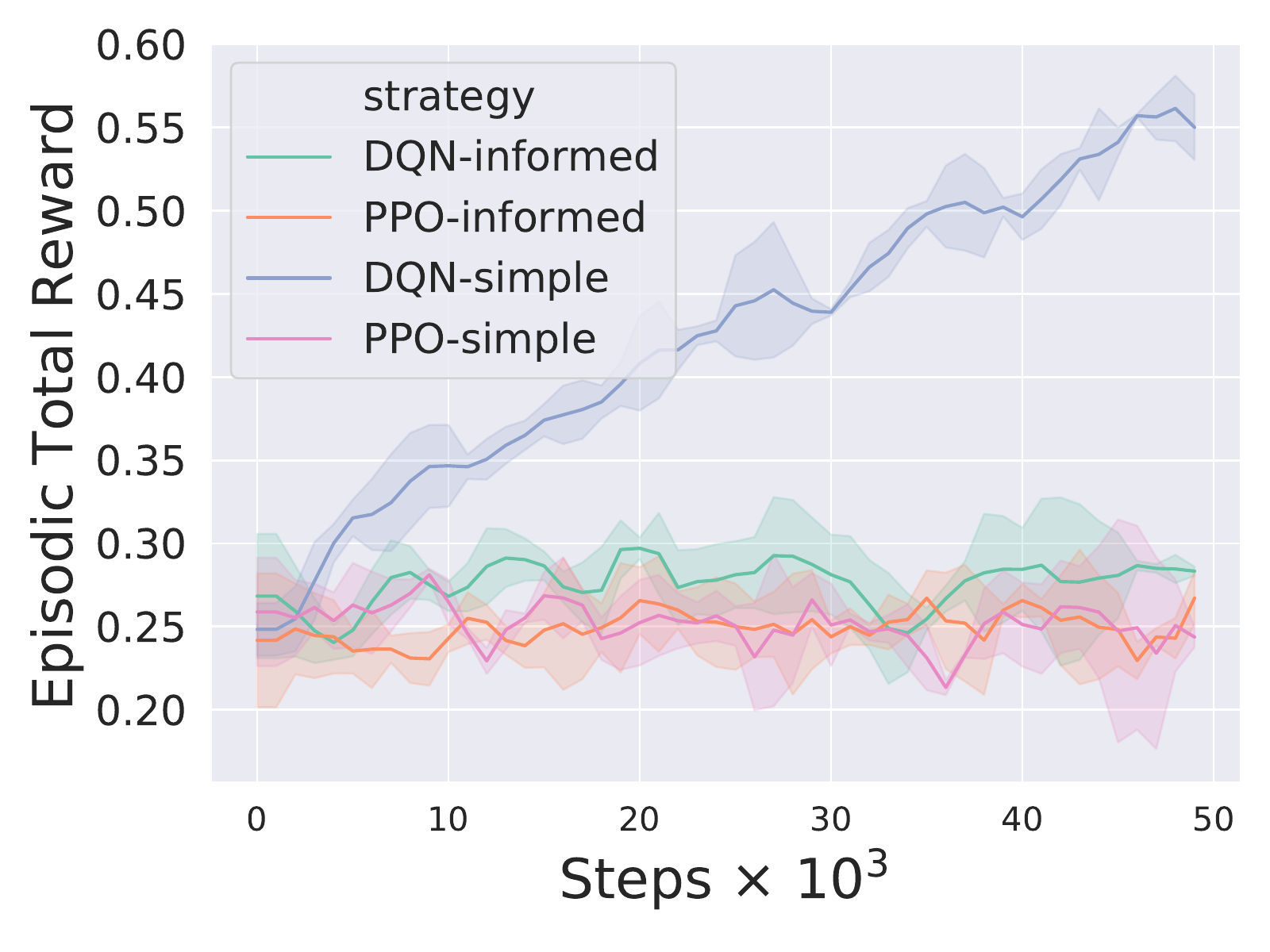}}
	
	\caption{\textbf{Learning Curves}: Averaged learning curves of PPO and DQN agents with variations. (a),(d) learning curves for sequence tagging with NER and POS tags. (b),(e) MLC with Reuters and AAPD datasets; (c), (f) QA tasks with QASC and AIRC datasets}
	\label{fig:learning_curves}
\end{figure*}
To demonstrate the usage of our toolkit, we run the provided environments by plugging-in with benchmark datasets and train Deep Q Networks (DQN) and Proximal Policy Optimization (PPO) agents using stable-baselines \cite{stable-baselines}. Each experiment is repeated $5$ times to obtain average learning curves. All our experiments are run in a batch setting (train data points are added to the environments) and evaluated on test split.

\paragraph{Sequence Tagging}We chose CONLL \cite{sang2003introduction} and UDPOS \cite{silveira2014gold} as datasets for ST environments. We vary the observation featurizer with fasttext \cite{joulin2016bag} and bytepair \cite{heinzerling2018bpemb} embeddings. For each of these settings, we train DQN and PPO agents.
For DQN, we chose a feedforward multi-layer perceptron (MLP) with two hidden layers, each consisting of 100 neurons. A discount factor of $0.99$ is used. The experiences are stored in a replay buffer and a batch size of $32$ is used for sampling. Additionally, Double-Q learning is used. For PPO, we chose the MLP policy with the same number of hidden layers and neurons. The other settings for PPO such as discount factor, batch size, clip parameter, entropy co-efficient are set to $0.99$, $64$, $0.2$ and $0.0$ respectively. Both PPO and DQN are trained with a learning rate of $0.0005$.

\paragraph{Question Answering}For the QA environment, we test it on QASC \cite{khot2019qasc} and AIRC \cite{allenai:arc} datasets. The QA experiments are run with two featurizers: \textit{simple} and \textit{informed} featurizers (see observation featurizers in Appendix \ref{appendix: default} for details). The Q network in DQN and policy network in PPO consists of two hidden layers, each consisting of $64$ neurons. Both are trained with a learning rate of $0.0001$. Other parameters are set to the same values specified in ST experiments.

\paragraph{Multi-label Classification} For the MLC environment, we picked Reuters \cite{reuters} and AAPD \cite{YangCOLING2018} datasets. Like in ST, we vary the observation featurizer with fasttext \cite{joulin2016bag} and bytepair \cite{heinzerling2018bpemb} embeddings. For this task, we chose Q-network and policy network to consist of two hidden layers, each with $200$ neurons and they are trained with a learning rate of $0.001$. Other settings are the same as specified in ST experiments.

\paragraph{Results and Discussion}Figure \ref{fig:learning_curves} shows learning curves for the described agents and Table \ref{tbl:scores} summarizes their generalization performance. Further, in Appendix \ref{appendix:outputs}, we present the predictions (actions) of trained agents. For each agent and its variation, we first evaluate them on validation set \textit{dev} and select one model based on the validation performance. The selected models are then evaluated on the hold-out \textit{test} set and their scores are summarized below in the Table \ref{tbl:scores}.
We report micro F1-scores for ST and MLC, and for QA, accuracy is reported. It is observed that both DQN and PPO agents are able to solve ST and MLC tasks with a very good performance and PPO performs better than DQN agents in most cases.

\begin{table*}[t!]
\centering
\setlength{\tabcolsep}{6pt} 
\renewcommand{\arraystretch}{1.2} 
\resizebox{0.9\columnwidth}{!}{%
\begin{tabular}{@{}cccccccc@{}}
\toprule
  \multicolumn{2}{c}{\textbf{Model/Task}} &
  \multicolumn{2}{c}{\textbf{Sequence Tagging}} &
  \multicolumn{2}{c}{\textbf{Multi-Label}} &
  \multicolumn{2}{c}{\textbf{Question Answering}} \\
  \cmidrule(lr){1-2}\cmidrule(lr){3-4}\cmidrule(lr){5-6} \cmidrule(lr){7-8}
 & Embedding & 
  CONLL &
  UDPOS &
  Reuters &
  AAPD &
  QASC &
  AIRC \\ 
 \cmidrule(lr){2-2} \cmidrule(lr){3-3}\cmidrule(lr){4-4}\cmidrule(lr){5-5}\cmidrule(lr){6-6}\cmidrule(lr){7-7}\cmidrule(lr){8-8}
\multirow{2}{*}{DQN} & Fasttext & 0.92 & 0.69 & 0.72 & 0.35 & \textbf{0.49} & \textbf{0.35} \\ 
 & Byte-Pair & 0.91 & 0.58 & 0.65 & 0.26 & - & - \\ 
 \midrule
\multirow{2}{*}{PPO} & Fasttext & \textbf{0.93} & \textbf{0.77} & \textbf{0.76} & \textbf{0.61} & 0.14 & 0.24\\ 
 & Byte-Pair & 0.91 & 0.65 & 0.70 & 0.49 & - & - \\ 

\bottomrule
\end{tabular}
}
\caption{\textbf{Test Performance}: Summary of performance of selected DQN and PPO agents on test set. For ST and MLC, micro F1-scores are reported and for QA, accuracy is reported}
\label{tbl:scores}
\end{table*}

However, QA tasks are difficult to solve in general and require more complex representations (eg. BERT \cite{devlin2018bert}) which would improve the results for QA experiments. From Figure \ref{fig:learning_curves} and Table \ref{tbl:scores}, it is seen that the agents seem to memorize the training questions resulting in poor generalization. Nevertheless, all of these results, indicate that RL agents can be indeed trained to perform NLP tasks using our toolkit.

\section{Conclusion}

In this work, we presented and demonstrated the usage of NLPGym for the application of DRL on solving NLP tasks. The initial release of this toolkit contains environments for three standard tasks that are ready-to-use with default components and datasets.  The results presented here act as simple baselines to promote the research and benchmarking of RL in NLP settings.  Moreover, there is a clear direction for future work: to release environments for other NLP tasks such as text summarization, generation, and translation.  We believe that NLPGym becomes a standard toolkit for testing agents for learning language and understanding.

\newpage

\appendix                                     
\section{Appendix}

\subsection{Default Featurizers} \label{appendix: default}

The toolkit provides its users with simple observation featurizers based on pre-trained word embeddings to get started with the environments without much setup.  They are easily extendable and replaceable with custom components (see \ref{appendix:extensibility})

\paragraph{Sequence tagging (ST):} \label{seq_tagging}  In this environment, observation at any time step is a word/token and a tagged label of the previous word. The word is vectorized into one of the pre-trained word embeddings such as fasttext \cite{joulin2016bag}, byte-pair \cite{heinzerling2018bpemb} and flair embeddings \cite{akbik-etal-2018-contextual}. On the contrary, the tagged label is converted to a one-hot encoded representation since the label vocabulary is known. In the end, the observation vector is simply a concatenation of the word vector and the label vector.

\paragraph{Multi-Label Classification (MLC):} The observation at any time step in this environment is a sentence and a generated label sequence up to the time step. The sentence is converted to a fixed-length representation by pooling over its corresponding word embeddings (pre-trained embeddings).  The label sequence is
converted to a Bag-of-Words representation (BoW).  To obtain the observation vector, the representations of these two are concatenated together.

\paragraph{Question-answering (QA):} For the QA environment, the observation is a triplet consisting of question, choice and facts. The toolkit offers two types of featurizers: \textit{simple} and \textit{informed}. The \textit{simple} featurizer converts the observation into a concatenated vector of sentence representations of the question, choice and facts.  On the contrary, \textit{informed} featurizer leverages the fact that the correct choice should be semantically similar to the question and the given facts than other choices. To this end, the observation is simply a 2-D vector consisting of cosine similarity between choice-question and choice-facts.

\subsection{Agent's Prediction} \label{appendix:outputs}

This section presents the predictions (actions) of trained agents on the given tasks by applying them on the hold-out set. The best PPO agent from each task has been selected for this introspection as it performed well in most experiments.

\subsubsection{Sequence Tagging}

We present the results of samples, predictions, true labels and their total reward for two datasets: CONLL and UDPOS, separately.

\paragraph{CONLL Dataset:}

\begin{minted}[fontsize=\small]{text}
--------------------------------------------------------------------------------
"text": "Justin suffered a sprained right shoulder in the third quarter 
         and did not return ."
"true_label": ["PER", "O", "O", "O", "O", "O", "O", "O", "O", "O", "O",
               "O", "O", "O", "O"],
"predicted_label": ["PER", "O", "O", "O", "O", "O", "O", "O", "O", "O", 
                    "O", "O", "O", "O", "O"],
"total_reward": 1.0
--------------------------------------------------------------------------------
"text": "Ricky Watters , who leads the NFC in rushing , left the game after 
         getting kneed to the helmet after gaining 33 yards on seven carries ."
"true_label": ["PER", "PER", "O", "O", "O", "O", "O", "O", "O", "O", "O", "O", 
               "O", "O", "O", "O", "O", "O", "O", "O", "O", "O", "O", "O", "O", 
               "O", "O"],
"predicted_label": ["PER", "PER", "O", "O", "O", "O", "ORG", "O", "O", "O", "O", 
                    "O", "O", "O", "O", "O", "O", "O", "O", "O", "O", "O", "O", 
                    "O", "O", "O", "O"],
"total_reward": 0.9629629629629629
--------------------------------------------------------------------------------
"text": "SACRAMENTO 6 12 .333 8",
"true_label": ["ORG", "O", "O", "O", "O"],
"predicted_label": ["LOC", "O", "O", "O", "O"],
"total_reward": 0.8000000000000002
--------------------------------------------------------------------------------
 "text": "Tampico port authorities said fishing restrictions were in place in 
          an area adjacent to the port because of a geophysical study being 
          carried out in deep   waters of the region from the ship Kenda ."
"true_label": ["LOC", "O", "O", "O", "O", "O", "O", "O", "O", "O", "O", "O", 
               "O", "O", "O", "O", "O", "O", "O", "O", "O", "O", "O", "O", "O", 
               "O", "O", "O", "O", "O", "O", "O", "O", "MISC", "O"],
"predicted_label": ["ORG", "O", "O", "O", "O", "O", "O", "O", "O", "O", "O", 
                    "O", "O", "O", "O", "O", "O", "O", "O", "O", "O", "O", 
                    "O", "O", "O",  "O", "O", "O", "O", "O", "O", "O", "O", 
                    "PER", "O"],
"total_reward": 0.9428571428571428
--------------------------------------------------------------------------------
"text": "BEIJING 1996-12-06",
"true_label": ["LOC", "O"],
"predicted_label": ["LOC", "O"],
"total_reward": 1.0
--------------------------------------------------------------------------------
\end{minted}

\newpage

\paragraph{UDPOS Dataset:}

\begin{minted}[fontsize=\small]{text}
--------------------------------------------------------------------------------
"text": "Wei Ligang , a rising star of modern art in China , just had an 
         exhibition in mid-July , 2005 in Hong Kong .",
"true_label": ["PROPN", "PROPN", "PUNCT", "DET", "VERB", "NOUN", "ADP", "ADJ", 
               "NOUN", "ADP", "PROPN", "PUNCT", "ADV", "VERB", "DET", "NOUN", 
               "ADP", "PROPN", "PUNCT", "NUM", "ADP", "PROPN", "PROPN", "PUNCT"],
"predicted_label": ["PROPN", "PROPN", "PUNCT", "DET", "VERB", "NOUN", "ADP", "ADJ", 
                    "NOUN", "ADP", "PROPN", "PUNCT", "ADV", "VERB", "DET", "NOUN", 
                    "ADP", "PROPN", "PUNCT", "NUM", "ADP", "PROPN", "PROPN", 
                    "PUNCT"],
"total_reward": 1.0
--------------------------------------------------------------------------------
"text": "What if Google Morphed Into GoogleOS ?",
"true_label": ["PRON", "SCONJ", "PROPN", "VERB", "ADP", "PROPN", "PUNCT"]
"predicted_label": ["PRON", "CCONJ", "PROPN", "VERB", "ADP", "PROPN", "PUNCT"]
"total_reward": 0.8571428571428571
--------------------------------------------------------------------------------
"text": "They know that the American advent implies for them a demotion , 
        and an elevation of the Shiites and Kurds , and 
        they refuse to go quietly .",
"true_label": ["PRON", "VERB", "SCONJ", "DET", "ADJ", "NOUN", "VERB", "ADP", 
               "PRON", "DET", "NOUN", "PUNCT", "CCONJ", "DET", "NOUN", "ADP", 
               "DET", "PROPN", "CCONJ", "PROPN", "PUNCT", "CCONJ", 
               "PRON", "VERB", "PART", "VERB", "ADV", "PUNCT"]
"predicted_label": ["PRON", "VERB", "PRON", "DET", "ADJ", "NOUN", "VERB", "ADP", 
                    "PRON", "DET", "NOUN", "PUNCT", "CCONJ", "DET", "NOUN", 
                    "ADP", "DET", "PROPN", "CCONJ", "PROPN", "PUNCT", "CCONJ", 
                    "PRON", "VERB", "ADP", "VERB", "ADV", "PUNCT"],

"total_reward": 0.9285714285714286
--------------------------------------------------------------------------------
"text": "...",
"true_label": ["SYM"]
"predicted_label": ["PUNCT"]
"total_reward": 0.0
--------------------------------------------------------------------------------
 "text": "Can police trace a cell phone even if it is switched off ?"
 "true_label": ["AUX", "NOUN", "VERB", "DET", "NOUN", "NOUN", "ADV", 
                "SCONJ", "PRON", "AUX", "VERB", "ADP", "PUNCT"]
 "predicted_label": ["AUX", "ADJ", "NOUN", "DET", "NOUN", "NOUN", 
                     "ADV", "CCONJ", "PRON", "AUX", "VERB", "ADP", "PUNCT"]
 "total_reward": 0.7692307692307693
 --------------------------------------------------------------------------------
\end{minted}

\newpage

\subsubsection{Multi-Label Classification}

\paragraph{Reuters Dataset}

\begin{minted}[fontsize=\small]{text}
--------------------------------------------------------------------------------
"text": "FED EXPECTED TO ADD RESERVES VIA CUSTOMER RPS\n  
         The Federal Reserve is expected to enter\n  
         the government securities market to add reserves via customer\n  
         repurchase agreements, economists said.\n      
         They expected the amount to total around 1.5 billion to two\n  
         billion dlrs.\n      
         Economists added that the low rate on federal funds\n  
         indicates the Fed is unlikely to add funds agressively through\n  
         overnight system repurchases, unless it feels the need to calm\n  
         volatile financial markets.\n      
         Federal funds were trading at 7-1/8 pct, down from\n  
         yesterday's average of 7.61 pct.\n  \n\n",
"true_label": ["interest", "money-fx"],
"predicted_label": ["money-fx", "interest"],
"total_reward": 1.0
--------------------------------------------------------------------------------
 "text": "NERCI &lt;NER> UNIT CLOSES OIL/GAS ACQUISITION\n  
          Nerco Inc said its oil and gas\n  
          unit closed the acquisition of a 47 pct working interest in the\n  
          Broussard oil and gas field from &lt;Davis Oil Co> for about 22.5\n  
          mln dlrs in cash.\n      
          Nerco said it estimates the field's total proved developed\n  
          and undeveloped reserves at 24 billion cubic feet, or\n  
          equivalent, of natural gas, which more than doubles the\n  
          company's previous reserves.\n      
          The field is located in southern Louisiana.\n  \n\n",
"true_label": ["acq", "crude", "nat-gas"],
"predicted_label": ["crude"],
"total_reward": 0.5
--------------------------------------------------------------------------------
 "text": "SOFTWARE SERVICES OF AMERICA INC &lt;SSOA.O> NET\n  
          3rd qtr Feb 28\n      
          Shr profit 14 cts vs loss four cts\n      
          Net profit 311,994 vs loss 66,858\n      
          Revs 2,229,273 vs 645,753\n      
          Nine mths\n      
          Shr profit 51 cts vs profit two cts\n      
          Net profit 1,126,673 vs profit 42,718\n      
          Revs 7,277,340 vs 1,378,372\n  \n\n",
"true_label": ["earn"],
"predicted_label": ["earn"],
"total_reward": 1.0
--------------------------------------------------------------------------------
 "text": "PORTUGUESE CONSUMER PRICES UP 1.4 PCT IN MARCH\n  
          Portugal's consumer prices rose 1.4 pct\n  
          in March after a one pct increase in February and a 1.2 pct\n  
          rise in March 1986, the National Statistics Institute said.\n      
          The consumer price index (base 100 for 1976) rose to 772.0\n  
          from 761.3 in February and compared with 703.4 in March 1986.\n      
          This gave a year-on-year March inflation rate of 9.8 pct\n  
          against 9.5 pct in February and 12.2 pct in March 1986.\n      
          Measured as an annual average rate, inflation in March was\n  
          10.9 pct compared with 11.1 pct in February. The government\n  
          forecasts annual inflation of about eight pct this year.\n  \n\n",
 "true_label": ["cpi"],
 "predicted_label": ["money-supply"],
 "total_reward": 0.0
 --------------------------------------------------------------------------------
\end{minted}

\newpage

\paragraph{AAPD Dataset}
\begin{minted}[fontsize=\small]{text}
--------------------------------------------------------------------------------
"text": "a relay channel is one in which a source and destination use an 
        intermediate relay station in order to improve communication rates 
        we propose the study of relay channels with classical inputs and 
        quantum outputs and prove that a partial decode and forward strategy 
        is achievable we divide the channel uses into many blocks and build 
        codes in a randomized , block markov manner within each block the relay 
        performs a standard holevo schumacher westmoreland quantum measurement 
        on each block in order to decode part of the source 's message and then 
        forwards this partial message in the next block the destination performs 
        a novel sliding window quantum measurement on two adjacent blocks in 
        order to decode the source 's message this strategy achieves non trivial
        rates for classical communication over a quantum relay channel",
"true_label": ["quant-ph", "cs.IT", "math.IT"],
"predicted_label": ["cs.IT", "math.IT"],
"total_reward": 0.8
--------------------------------------------------------------------------------
"text": "recently , social phenomena have received a lot of attention not only 
         from social scientists , but also from physicists , mathematicians and 
         computer scientists , in the emerging interdisciplinary field of complex 
         system science opinion dynamics is one of the processes studied , 
         since opinions are the drivers of human behaviour , and play a crucial 
         role in many global challenges that our complex world and societies are 
         facing global financial crises , global pandemics , growth of cities , 
         urbanisation and migration patterns , and last but not least important , 
         climate change and environmental sustainability and protection opinion 
         formation is a complex process affected by the interplay of different 
         elements , including the individual predisposition , the influence of 
         positive and negative peer interaction \\( social networks playing 
         a crucial role in this respect \\) , the information each individual is 
         exposed to ,and many others several models inspired from those in use in 
         physics have been developed to encompass many of these elements , and to 
         allow for the identification of the mechanisms involved in the opinion 
         formation process and the understanding of their role , with the 
         practical aim of simulating opinion formation and spreading under various 
         conditions these modelling schemes range from binary simple models such 
         as the voter model, to multi dimensional continuous approaches here , 
         we provide a review of recent methods , focusing on models employing 
         both peer interaction and external information , and emphasising the 
         role that less studied mechanisms , such as disagreement , has in 
         driving the opinion dynamics",
"true_label": ["physics.soc-ph", "cs.SI"],
"predicted_label": ["physics.soc-ph", "cs.SI"],
--------------------------------------------------------------------------------
"text": "we propose a new platform for implementing secure wireless ad hoc 
         networks our proposal is based on a modular architecture , 
         with the software stack constructed directly on the ethernet layer 
         within our platform we use a new security protocol that we designed 
         to ensure mutual authentication between nodes and a secure key exchange
         the correctness of the proposed security protocol is 
         ensured by guttman 's authentication tests",
"true_label": ["cs.CR", "cs.NI"],
"predicted_label": ["cs.CR", "cs.NI"],
"total_reward": 1.0
--------------------------------------------------------------------------------
\end{minted}

\newpage

\subsubsection{Question Answering}

\paragraph{QASC Dataset}
\begin{minted}[fontsize=\small]{text}
--------------------------------------------------------------------------------
"question": "What can increase the chances of flooding?",
"facts": ["if weather is stormy then there is a greater chance of rain", 
          "Rain is good, but lots of rain causes destructive flooding."],
"choices": {
    "A": "filter feeders",
    "B": "low tide",
    "C": "permeable walls",
    "D": "fortifying existing levees",
    "E": "higher corn prices",
    "F": "stormy weather",
    "G": "Being over land",
    "H": "feedback mechanisms"
},
"true_label": "F",
"predicted_label": "F",
"total_reward": 1.0
--------------------------------------------------------------------------------
"question": "How can animals like cyptosporidium be classified?",
"facts": ["Protozoa can be classified on the basis of how they move.", 
          "Cryptosporidium parvum is the hardest protozoa to kill."],
"choices": {
    "A": "prokaryotic cells",
    "B": "holding nutrients",
    "C": "Laboratory",
    "D": "how they move",
    "E": "eukaryotic cells",
    "F": "coelenterates",
    "G": "melanin content",
    "H": "angiosperm"
},
"true_label": "D",
"predicted_label": "A"
--------------------------------------------------------------------------------
"question": "What can magnets be used to do?",
"facts": ["a compass is used for determining direction", 
          "Magnets are used in compasses."],
"choices": {
    "A": "capture prey",
    "B": "moving over land",
    "C": "feedback mechanisms",
    "D": "Destroy magnets",
    "E": "reproduce",
    "F": "help other species benefit",
    "G": "Direct a play",
    "H": "Determine direction"
},
"true_label": "H",
"predicted_label": "H",
"total_reward": 1.0
--------------------------------------------------------------------------------
\end{minted}

\newpage

\subsection{Demo Scripts}

\subsubsection{Sample Interaction with QA environment}

\begin{minted}[fontsize=\small]{python}
from nlp_gym.data_pools.custom_question_answering_pools import QASC
from nlp_gym.envs.question_answering.env import QAEnv

# data pool
pool = QASC.prepare("train")

# custom answering env
env = QAEnv()
for sample, weight in pool:
    env.add_sample(sample)

# play an episode
done = False
state = env.reset()
total_reward = 0
while not done:
    action = env.action_space.sample()
    state, reward, done, info = env.step(action)
    total_reward += reward
    env.render()
    print(f"Action: {env.action_space.ix_to_action(action)}")
print(f"Total reward: {total_reward}")
\end{minted}

\begin{minted}{pycon}
Step 0
Question: Machines can use gasoline to do what?
Fact: a gasoline lawn mower converts gasoline into motion
Fact: Machines mow down forests much as a lawn mower cuts grass.
Choice A: energy
Action: CONTINUE
Step 1
Question: Machines can use gasoline to do what?
Fact: a gasoline lawn mower converts gasoline into motion
Fact: Machines mow down forests much as a lawn mower cuts grass.
Choice A: energy
Fact: a gasoline lawn mower converts gasoline into motion
Fact: Machines mow down forests much as a lawn mower cuts grass.
Choice B: destroy matter
Action: ANSWER
Total reward: 0.0
\end{minted}

\newpage

\subsubsection{Training PPO/DQN Agents using baselines} \label{appendix:ppo_dqn_sample}

In the code snippet below, we present the code for instantiating sequence tagging environment with CONLL dataset and train it using PPO and DQN agents from stable-baselines \cite{stable-baselines}

\begin{minted}[fontsize=\small]{python}
from nlp_gym.data_pools.custom_seq_tagging_pools import UDPosTagggingPool
from nlp_gym.envs.seq_tagging.env import SeqTagEnv
from nlp_gym.envs.seq_tagging.reward import EntityF1Score
from stable_baselines.deepq.policies import MlpPolicy as DQNPolicy
from stable_baselines import DQN
from stable_baselines.common.env_checker import check_env
from rich import print


# data pool
data_pool = UDPosTagggingPool.prepare(split="train")

# reward function
reward_fn = EntityF1Score(dense=True, average="micro")

# seq tag env
env = SeqTagEnv(data_pool.labels(), reward_function=reward_fn)
for sample, weight in data_pool:
    env.add_sample(sample, weight)

# check the environment
check_env(env, warn=True)

# train a MLP Policy using DQN
model = DQN(env=env, policy=DQNPolicy, gamma=0.99, batch_size=32, learning_rate=5e-4,
            double_q=True, exploration_fraction=0.1,
            prioritized_replay=False, policy_kwargs={"layers": [100, 100]},
            verbose=1)
model.learn(total_timesteps=int(1e+4))
\end{minted}

\newpage

\subsubsection{Online Learning} \label{appendix: online}
We illustrate below the usage of the environment in an online setting. In each iteration, one sample (data point) is added to the environment for sampling. This allows us to develop interactive algorithms (for instance, active learning systems) that involve human interaction. A complete run of such an algorithm corresponds to precisely one pass over the entire dataset.

\begin{minted}[fontsize=\scriptsize]{python}


from nlp_gym.envs.seq_tagging.env import SeqTagEnv
from nlp_gym.data_pools.custom_seq_tagging_pools import UDPosTagggingPool
from stable_baselines.common.policies import MlpPolicy
from stable_baselines import PPO1
from nlp_gym.envs.seq_tagging.reward import EntityF1Score
from nlp_gym.envs.seq_tagging.featurizer import DefaultFeaturizerForSeqTagging
from nlp_gym.metrics.seq_tag import EntityScores
import tqdm


def predict(model, sample):
    done = False
    obs = env.reset(sample)
    predicted_label = []
    while not done:
        action, _ = model.predict(obs)
        obs, _, done, _ = env.step(action)
        predicted_label.append(env.action_space.ix_to_action(action))
    return predicted_label


# data pool
data_pool = UDPosTagggingPool.prepare(split="train")

# reward function
reward_fn = EntityF1Score(dense=True, average="micro")

# seq tag env
env = SeqTagEnv(data_pool.labels(), reward_function=reward_fn)

# observation featurizer
feat = DefaultFeaturizerForSeqTagging(env.action_space, embedding_type="fasttext")
env.set_featurizer(feat)

# PPO model
model = PPO1(MlpPolicy, env, verbose=0)


# train loop that goes over each sample only once
running_match_score = 0
for ix, (sample, _) in enumerate(tqdm.tqdm(data_pool)):

    # run the sample through the model and get predicted label
    predicted_label = predict(model, sample)

    # after few epochs, predicted_label can be used as pre-annotated input
    # then the user can just correct it
    # to reduce human efforts

    # get annotated label from user (just simulated for now)
    annotated_label = sample.oracle_label

    # match score
    match_ratio = EntityScores()(annotated_label, predicted_label)["f1"]
    running_match_score += match_ratio

    # add the new sample to the environment
    sample.oracle_label = annotated_label
    env.add_sample(sample)

    # train agent for few epochs
    model.learn(total_timesteps=1e+2)

    if (ix+1) % 50 == 0:
        print(f"Running match score {running_match_score/50}")
        running_match_score = 0.0


\end{minted}
\newpage

\subsection{Custom Components} \label{appendix:extensibility}

The environments are ready to use with default implementations for observation featurizers, reward functions and datasets. However, the toolkit is modular in a way that users can plug-in their own implementations of these components.  This gives the flexibility to implement some of the components to have trainable components (e.g. observation featurizer) that can be optimized either end-to-end or by pre-training.

\subsubsection{Datasets} 
For instance, to create a custom sequence tagging dataset, users must provide a list of samples (data points) and possible labels (that correspond to agent actions). These samples should be instances of the \mintinline{python}{Sample} data class as shown below.

\begin{minted}[fontsize=\small]{python}

class Sample:
    """
    Dataclass for holding datapoints
    
    Attributes:
        input_text - textual input
        oracle_label - true label for the given data point
    """
    input_text: str
    oracle_label: List[str]
\end{minted}

With this setup, the creation of a custom dataset is straightforward, as shown below:

\begin{minted}[fontsize=\small]{python}

# custom dataset
custom_dataset = SeqTaggingPool(samples, possible_labels=[""])
\end{minted}

\subsubsection{Reward Function} Users can define their own custom reward function by sub-classing \mintinline{python}{RewardFunction} which is a \mintinline{python}{Callable} that takes \mintinline{python}{BaseObservation}, \mintinline{python}{current action}, \mintinline{python}{targets} and returns a scalar reward.

\begin{minted}[fontsize=\small]{python}
from nlp_gym.envs.common.observation import BaseObservation
from nlp_gym.envs.common.reward import RewardFunction

class MyRewardFunction(RewardFunction)
    def __call__(self, observation: BaseObservation, action: str, targets: List[str]) -> float:
        """
        My custom reward function
        Args:
            observation (BaseObservation): current observation at t
            action (str): current action at t
            targets (List[str]): targets of the current sample

        Returns:
            - a scalar reward
        """
        pass
        
\end{minted}

\newpage

\subsubsection{Observation Featurizer} Similarly, observation featurizer can have its custom implementation by sub-classing it from \mintinline{python}{BaseObservationFeaturizer} as shown below. Each featurizer must implement \mintinline{python}{featurize()} and \mintinline{python}{get_observation_dim()}. For instance, for the sequence tagging environment, the featurizer must additionally implement \mintinline{python}{init_on_reset()} which is called by the environment on \mintinline{python}{reset()}.

\begin{minted}[fontsize=\small]{python}


class BaseObservationFeaturizer(ABC):

    @abstractmethod
    def featurize(self, observation: BaseObservation) -> torch.Tensor:
        raise NotImplementedError

    def get_observation_dim(self) -> int:
        """
        Returns the observation dim
        """
        return self.get_input_dim() + self.get_context_dim()

# for sequence tagging environment
class ObservationFeaturizer(BaseObservationFeaturizer):

    @abstractmethod
    def init_on_reset(self, input_text: Union[List[str], str]):
        """
        Takes an input text (sentence) or list of token strings and featurizes it or prepares it
        This function would be called in env.reset()
        """
        raise NotImplementedError


\end{minted}

\end{document}